\documentclass[runningheads]{llncs}
\usepackage[T1]{fontenc}
\usepackage{graphicx}
\usepackage{booktabs}
\usepackage[misc]{ifsym}

\usepackage{subcaption} % For subfigures
\usepackage{xcolor}
\usepackage{dsfont}
\usepackage{amsmath}
\usepackage{amssymb}
\usepackage{hyperref}
\usepackage{cleveref}
\usepackage{thmtools}
\usepackage{mathtools}
\usepackage{algorithm}
\usepackage{algorithmic}
\usepackage{thm-restate}
\usepackage{subcaption}
\usepackage{multirow}
\usepackage{lineno}

\usepackage{optidef}

\begin{document}
\title{DISCO: An End-to-End Bandit Framework for Personalised Discount Allocation}

\titlerunning{DISCO: A Bandit Framework for Personalised Discount Allocation}
% \title{Recent Advances in Underwater Basket Weaving Under the Extreme Pressure of the Mariana Trench}

% \titlerunning{Underwater Basket Weaving Under Extreme Pressure}
% If the full title of your paper is short enough to also fit in the running head, you can omit the abbreviated paper title here. You can check as follows: if you comment out the \titlerunning line, something will appear in the header of all odd-numbered pages of your PDF from page 3 onward. This something is either the full title (in which case all is well), or the error message "Title Suppressed Due to Excessive Length". If this error message appears, you're going to want to provide an abbreviated title within the \titlerunning command, because if you won't do it, Springer will do it for you.

%N.B.: Author information (both in the \author{} and \authorrunning{} command) should only be present in the Camera-Ready Version of your paper. The version that you initially submit for review, ought to be double-blind. So, when initially submitting your paper, use:
%\author{Author information scrubbed for double-blind reviewing}
\author{Jason Shuo Zhang \thanks{Now at Tripadvisor.} \and
Benjamin Howson \thanks{Work done while an intern at ASOS.com.}  \and
Panayiota Savva \and \\
Eleanor Loh \Letter}

% You may leave out the orcidID information, if you want to.
% Use \corr to indicate the corresponding author. Note the spacing around the \corr command. Only one author can be the corresponding author.

%N.B.: comment out the \authorrunning{} command for the double-blind version of your paper submitted for review. Later, if your paper is accepted, use the command for the Camera-Ready Version.
% \authorrunning{A.L. Benjamin et al.}
\authorrunning{Jason Shuo Zhang et al.}
% If there are two authors, write 'A.L. Benjamin and C.C. Broadus Jr.'
% If there are more than two authors, '[...] et al.' is used.

\institute{ASOS.com, London, United Kingdom \email{jasonzhang.co@gmail.com} \and
Imperial College London, United Kingdom \email{b.howson20@imperial.ac.uk} \and
ASOS.com, London, United Kingdom \email{panayiota.savva@asos.com} \and
ASOS.com, London, United Kingdom \email{eleanor.loh@asos.com}}

\tocauthor{Jason Shuo Zhang, Benjamin Howson, Panayiota Savva, Eleanor Loh}

\toctitle{DISCO: An End-to-End Bandit Framework for Personalised Discount Allocation}

\maketitle              % typeset the header of the contribution

\begin{abstract}
Personalised discount codes provide a powerful mechanism for managing customer relationships and operational spend in e-commerce. Bandits are well suited for this product area, given the partial information nature of the problem, as well as the need for adaptation to the changing business environment. Here, we introduce DISCO, an end-to-end contextual bandit framework for personalised discount code allocation at ASOS.com. DISCO adapts the traditional Thompson Sampling algorithm by integrating it within an integer program, thereby allowing for operational cost control. Because bandit learning is often worse with high dimensional actions, we focused on building low dimensional action and context representations that were nonetheless capable of good accuracy, including in extrapolation. Additionally, we sought to build a model that preserved the traditional relationship between price and sales, in which customers increasing their purchasing in response to lower prices ("negative price elasticity"). These aims were achieved by using radial basis functions to represent the continuous (i.e. infinite armed) action space, in combination with context embeddings extracted from a neural network. These feature representations were used within a Thompson Sampling framework to facilitate exploration, and further integrated with an integer program to allocate discount codes across ASOS's customer base. These modelling decisions result in a reward model that (a) enables pooled learning across similar actions, (b) is highly accurate, including in extrapolation, and (c) preserves the expected negative price elasticity. Through offline analysis, we also show that DISCO is able to effectively enact exploration and improves its performance over time, despite being subject to the global constraint. Finally, we subjected DISCO to a rigorous online A/B test, and find that it achieves a significant improvement of >1\% in average basket value, relative to the legacy systems.

\end{abstract}

\section{Introduction}
\label{sec:intro}

The ability to tailor discounts to different customers is a major source of competitive efficiency for retailers. Scrutiny of each customer's preferences and behaviour can help businesses algorithmically target their policies for increased customer loyalty and engagement. While beneficial, personalised discounting can bring a set of technical challenges that need to be carefully addressed to ensure long-term effectiveness and control over operational spend.
% While beneficial, such personalized pricing algorithms can however be tricky to design. 
% \textcolor{green}{When designed carefully, personalised pricing algorithms can also give a retailer the ability to adapt their operations to different tactical goals, such as prioritising efficiency in economically challenging times or prioritising customer retention in markets where customer acquisition costs are high.} 

Price and discount optimization systems face a fundamental problem of \textbf{partial information} \cite{loh2022promotheus,hua2021markdowns,zhu2022modeling}: outcome information is observed only for the specific pricing decisions that have been enacted in the past, and the absence of counterfactual outcomes in the historical dataset can undermine the ability of standard supervised learning methods to accurately predict demand in response to different policy changes. This is similar to the decision making problem as formulated within the bandit framework: learning takes place under partial information (rather than full supervision), and attempts by practitioners to develop new policies can produce action sets that overlap poorly with historical data and are thus difficult to train and evaluate offline \cite{li2022map}. Although some businesses address partial information situations by explicitly collecting randomised data in a one-off exercise (i.e. instead of relying on biased observational data; e.g. \cite{Gupta2019linkedin,mehrotra2020bandit}), the ever-changing nature of the business environment can swiftly render these datasets obsolete. Within machine learning, \textbf{contextual bandit methods} offer a principled and effective way of tackling the technical challenges of partial information \cite{li2022map,BanditsBook,sutton2018reinforcement}. The key strength of the bandit framework lies in its ability to express uncertainty and enact strategic exploration of regions of the state-action space where uncertainty is high. Contextual bandits are used to make algorithmic interventions in a variety of online settings, such as news and product recommendation \cite{mussi2022pricing,li2010contextual,mcinerney2018explore,mehrotra2020bandit}. Compared to greedy approaches that maximise reward within each instance, bandits have been shown to collect a more diverse set of data \cite{han2021budget}, and active learning has been shown to outperform greedy approaches in various settings over time.

In the contextual bandit framework, features are used to predict rewards associated with different actions in various contexts, and these predicted rewards are then converted into actions via bandit algorithms like Thompson Sampling or UCB. A major challenge in building bandit systems lies in building the model's representations: such action and context representations must be sufficiently rich in order to accurately predict rewards, but the performance of bandit algorithms is also known to degrade as the dimensionality of the action set increases \cite{saito22largeaction,zhu2022pmlr_largeaction}. While effective solutions have been devised for discrete, low-dimensional action spaces, efficiently implementing bandits with \textit{continuous} action spaces remains an area of active research (e.g. \cite{zhu2022pmlr_largeaction,krishnamurthy20smoothing,majzoubi2020efficient,srinivas2012infotheor}). As a result, practitioners dealing with continuous action space often resort to discretizing (e.g. \cite{kleinberg2004discretize}), which leads to a high dimensional action sets and detracts from the model's ability to pool its learning across actions that are closely related to each other (e.g. neighbouring points on a continuous action space). Because pricing problems naturally involve a continuous action set, our focus here is on developing an action representation scheme that (a) preserves information about the adjacency of different actions, (b) combines low dimensionality with the high degree of expressive complexity that is necessary for accurate predictions \cite{riquelme2018bakeoff}. Additionally, we adapt the standard contextual bandits approach by embedding it within an optimization framework that allows for a high degree of operational control over the overall system's budgetary spend. Bandits algorithms are known to effectively manage the explore-exploit trade-off in the \textit{unconstrained} case where the learner is able to sample actions freely based on the agent's subjective uncertainty; adding further constraints to systems's behaviour by constraining the kinds of choices it can make has the potential to degrade the system's ability to learn over time. However, we show here that our bandit system is able to perform well even with the addition of these specific operational constraints, without which such systems would not be useable at all in many practical situations. Lastly, we aimed to construct a model that preserves the conventional inverse relationship between price and consumer demand, known as negative price elasticity. This characteristic is a fundamental assumption of models within this problem domain, and is a crucial indicator of model validity: pricing models that lack such negative price elasticity often suffer from lack of interpretability and poor generalizability, and, in our direct experience, such models tend to be incorrect when used within algorithmic decision making systems. Pricing practitioners across various industries have made significant efforts to develop models that exhibit this specific behaviour, such as employing specialized modelling structures \cite{mussi2022pricing,zhu2022modeling}, custom loss functions \cite{ye2018customized,shukla2019dynamic}, or econometric/causal inference techniques \cite{hua2021markdowns,banerjee2016dynamic,goldenberg2022personalizing}). The complexity of these endeavours underscores both the importance as well as the difficulty of negative price elasticity within this and many other problem domains.

\begin{figure}[t]
    \centering 
    \includegraphics[width=\textwidth]{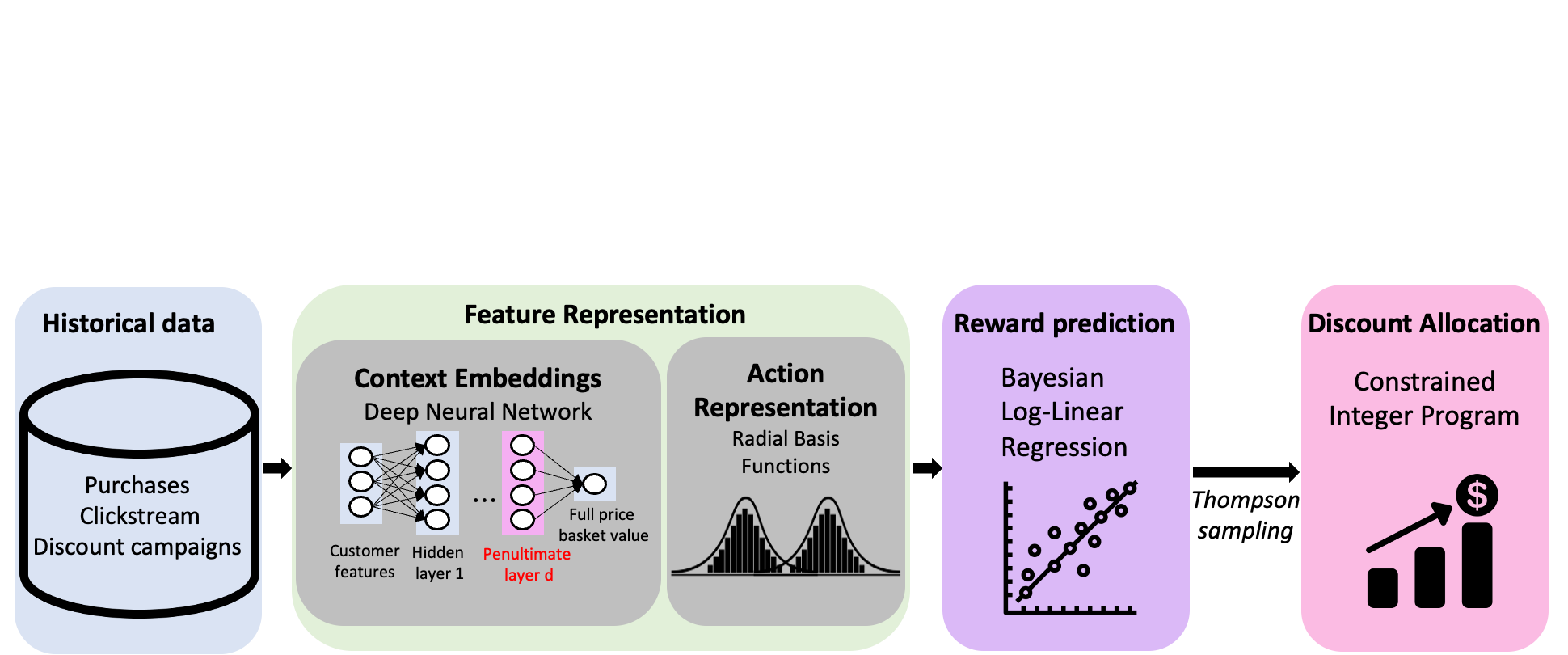}
    \caption{Overview of DISCO. DISCO uses low dimensional context embeddings (from a neural network) alongside radial basis functions that represent a continuous action space with low cardinality. These action representations enable pooled learning across similar actions. Features are used within a Bayesian log-linear regression to predict basket-level revenue (the reward signal). Constrained integer programming is then used to allocate discounts with operational control.}
    \label{figure:DISCO}
\end{figure}

In this paper, we introduce DISCO, a contextual bandit framework for allocating personalised discount codes at ASOS.com (Figure \ref{figure:DISCO}). We focus on providing practical solutions to key technical challenges, and outline a novel and effective way of (a) constructing performant bandit representations for continuous actions (b) integrating bandit methods with global constraints, in order to combine active learning with operational control. Specifically, we (i) encode our action space with radial basis functions, (ii) combine these representations with context embeddings generated from a neural network, (iii) use Thompson sampling to enact exploration, and (iv) embed our active learning model within a constrained integer program that allows the business to control the overall distribution of allocated discounts. The proposed action scheme maintains a low-dimensional representation to support more efficient bandit learning and allows our predictive model to achieve a high accuracy by enabling a high degree of expressive complexity. We show that this approach (i) supports shared learning between similar actions, (ii) maintains good predictive accuracy even when models encounter new actions (i.e. extrapolation), and (iii) produces demand curves that exhibit the expected negative price elasticity. We use simulations to demonstrate the superiority of active learning over greedy approaches over time, and also demonstrate that the addition of the integer program constraint incurs only a limited negative effect on the system's ability to enact active learning (relative to the more conventional unconstrained case). Finally, we validate our framework by subjecting it to a rigorous online test, where it outperforms legacy approaches to differentiated and undifferentiated discount code policies by >1\%.
% We also show that our algorithm is capable of both actively improving over time (i.e. benefiting from active learning), as well as adhering to business constraints that are widely applicable across e-commerce. Lastly, we demonstrate the efficacy of our framework via a rigorous online test, which outperformed the control code allocation approach by >3\%.
% Moreover, our algorithm demonstrates the capability for continuous improvement over time through active learning and adheres to widely applicable business constraints in the e-commerce domain. Finally, the efficacy of our framework is validated through a rigorous online 
% A/B test, where it outperforms the random code allocation approach by more than 3\%.

% \begin{figure}
%     \centering 
%     % \includegraphics[scale = 0.3]{figures/DISCO_small_color.pdf}
%     \includegraphics[scale = 0.2]{figures/DISCO_horizontal.png}
%     \caption{Overview of DISCO. DISCO uses low dimensional context embeddings (from a neural network) alongside radial basis functions that represent a continuous action space with low cardinality. These action representations enable pooled learning across similar actions. Features are used within a Bayesian log-linear regression to predict basket-level revenue (the reward signal). Constrained integer programming is then used to allocate discounts with operational control.}
%     \label{figure:DISCO}
% \end{figure}

\section{Problem formulation}
\label{sec:problem}

We aim to allocate different "\% off" discount codes across customers, to optimise downstream goals (e.g. maximise revenue). We refer to the expected \textit{full price} basket value $F_{t,i,a}$ of a given customer $i$ with discount $a$ in the $t$-th campaign:
\begin{equation}
\label{eq:fullpricemodel}
\small
\mathbb{E}\left[F_{t, i, a} \mid X_{t, i}, A_{t, i} = a\right] = g\left(X_{t, i}, a\right)
\end{equation}
where $X_{t, i}$ is contextual information, $A_{t, i}$ is the discount "\% off" given to the customer $i$ in campaign $t$ (note: $A_{t,i}=0.2$ indicates a "20\% off" discount code), and $g(.)$ refers to a mapping function between ($X_{t,i}\,,\,A_{t, i}$) and $F_{t, i, a}$. All discount codes are single-use, with specific expiry times (e.g. 1-31 days; we ignore expiry time in this paper). Full price basket values refer to the total currency value of checked-out baskets \textit{before} discounts are applied. Contextual bandits operate in rounds $t = 1,2...T$, and aim to actively balance the explore-exploit trade-off over time.  Within each round, the learner is presented with a batch of customers and their contexts $X_{t,i}$, and allocates a discount depth $a_{t,i} \in \mathcal{A}_{t} = \{a_1, a_2, \cdots, a_{K_t}\}$ for each customer. The learner then observes a batch of rewards (in this case, $F_{i,a} \forall i \in I$), and uses them to update model $g(.)$ for future inference.

Our decision to model \textit{full price} (vs discounted) basket values was based on the initial observation of monotonicity between discounts and \textit{full price} basket values (Figure \ref{figure:monotonicity} (left)): customers responded to deeper discounts by increasing the full price value of purchases, without necessarily leading to an increase in discounted basket value (which is computed $F_{t,i,a} * (1-A_{t,i})$). We also refer to "markdown cost", $C_{t,i,a} = F_{t,i,a} * A_{t,i}$, which measures the cost of applying discounts of a given level, and is commonly used in retail to constrain promotional activity \cite{Yao2015MarkdownOF,loh2022promotheus,wood21markdown}. A campaign's total cost is computed by aggregating markdown costs across all engaged customers.

\section{DISCO ARCHITECTURE}
\label{section: disco}

% DISCO (Figure \ref{figure:DISCO}) begins by using a neural net to extract customer embeddings that can be used for context representation in the bandit model. Then, a continuous action set is encoded at a lower dimensionality using radial basis functions. 

The contextual bandit formulation requires us to first build feature representations, $\psi: \mathcal{X} \times \mathcal{A} \rightarrow \mathbb{R}^{d}$, to encode the actions and contexts. DISCO (Figure \ref{figure:DISCO}) begins by transforming the continuous action set into a low-dimensional representation using radial basis functions. Then, a neural net is employed to extract customer embeddings, which serve as contextual representations. These action/context features are combined with a Bayesian log-linear regression model to predict customer-level \textit{full price} basket values as a function of discount depth. Lastly, an integer program is used to allocate discount depths across customers, subject to constraints specified by operational teams, and using likely customer-level rewards (generated via Thompson sampling) as an input.
% subject to constraints specified by operational teams, and using likely customer-level rewards (generated via Thompson sampling) as an input. 
\subsection{Action feature representation}\label{sec:rbf}

The natural action space consists of a continuous scale of discount depths. Although depth can be straightforwardly encoded as a continuous variable, this implies a linear relationship between depths and outcomes (due to the use of a linear model), with extensive feature engineering and functional form assumptions required to specify more realistic relationships. To overcome these limitations, we sought an alternative action encoding scheme, $\psi_2 \in \mathbb{R}^{d_2}$, that would be capable of generating low dimensional representations of the action space (similar to embeddings; \cite{saito22largeaction}). We prioritized low dimensionality in order to preserve the efficiency of bandit learning, which degrades as the cardinality of the action space increases \cite{BanditsBook}. 
% representations that (a) were low-dimensional, in order to limit the complexity of bandit learning \cite{BanditsBook} (b) would represent the relative similarity of different actions to each other, since discounts of similar in magnitude are likely to be positively correlated in their effect. 
We also avoided one-hot encoding (discretization) as it does not allow for information sharing, and increases the odds of limited support under offline evaluation \cite{saito22largeaction}. Instead, we use radial basis functions (RBFs) to encode the action-space. These functions measure the similarity between the selected basis locations and any given discount, and have the functional form:
\begin{equation}
\small
\psi_{2, z}\left(a \,\vert \mu_z, \alpha_z \right) = \exp\left(-\frac{\left(a - \mu_z\right)^{2}}{2 \alpha_z}\right)
\end{equation}
\noindent with $\psi_{2} = \{\psi_{2, z}\left(a \,\vert \mu_{z}, \alpha_{z} \right)\}_{z = 1}^{d_2} \in\mathbb{R}^{d_2}$. % \{\psi_{2, k}\left(a \,\vert \mu_{k}, \alpha_{k} \right)\}_{k = 1}^{K}$. 
% Figure \ref{figure: action feature-mapping} illustrates the use of $d_2 = 3$ three radial basis functions (at locations $0.25, 0.50, 0.75$) to represent the entire action space, as well as the effective number of times each action was selected by the algorithm when we have played $0.40, 0.60$ and $0.80$ 1K times each, illustrating that we gain information about actions close to the queried points, despite not playing them directly. 

\begin{figure*}[t]
    \centering
\begin{subfigure}[b]{0.32\textwidth}
    \centering
    \includegraphics[width=\textwidth]{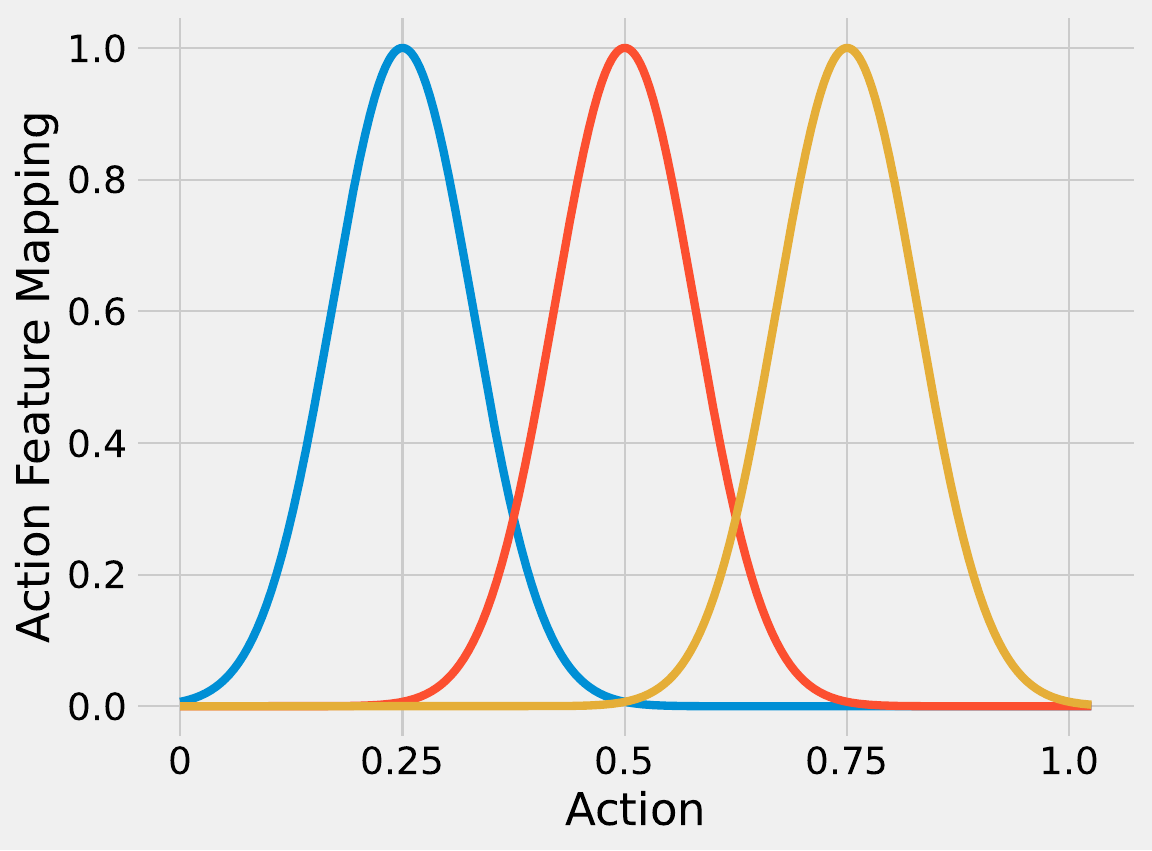} 
    % \caption{}
    \label{fig:rbf1}
\end{subfigure}
\hfill
\begin{subfigure}[b]{0.32\textwidth}
    \centering
    \includegraphics[width=\textwidth]{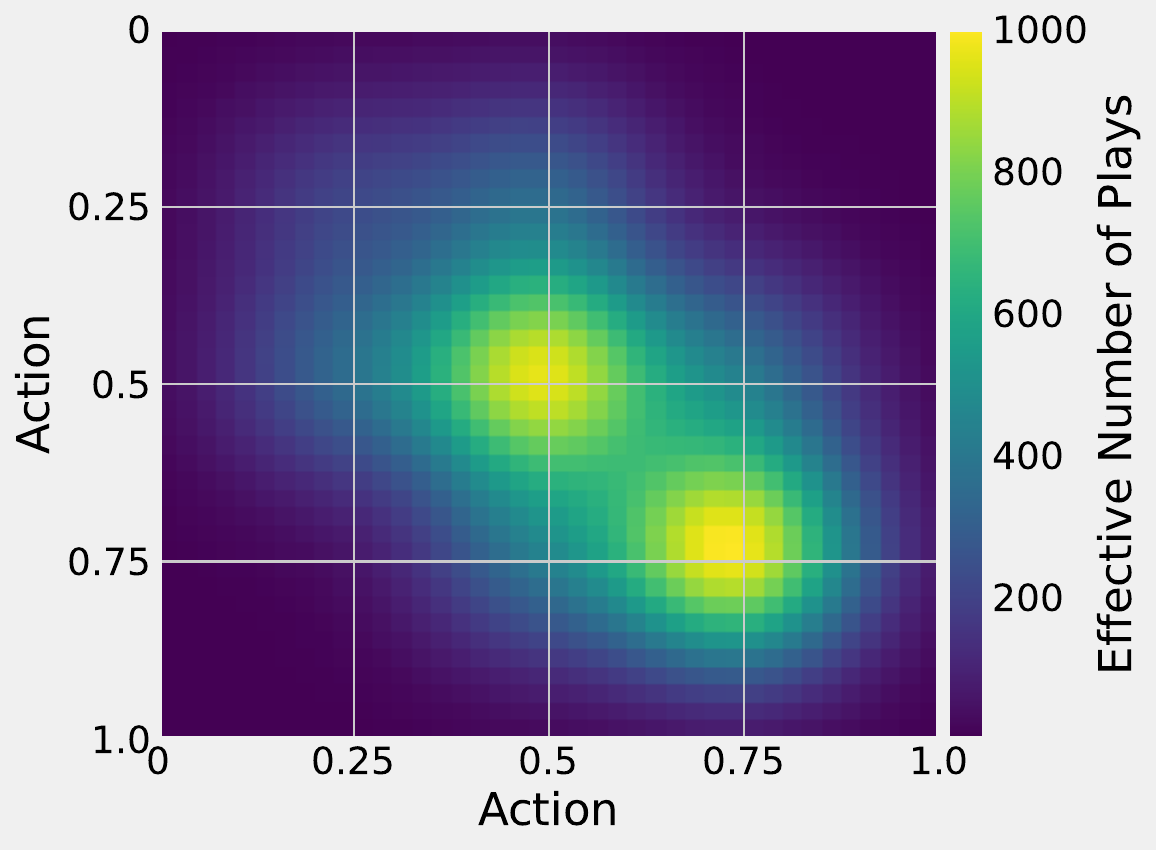} 
    % \caption{}
    \label{fig:rbf2}
\end{subfigure}
\hfill
\begin{subfigure}[b]{0.32\textwidth}
    \centering
    \includegraphics[width=\textwidth]{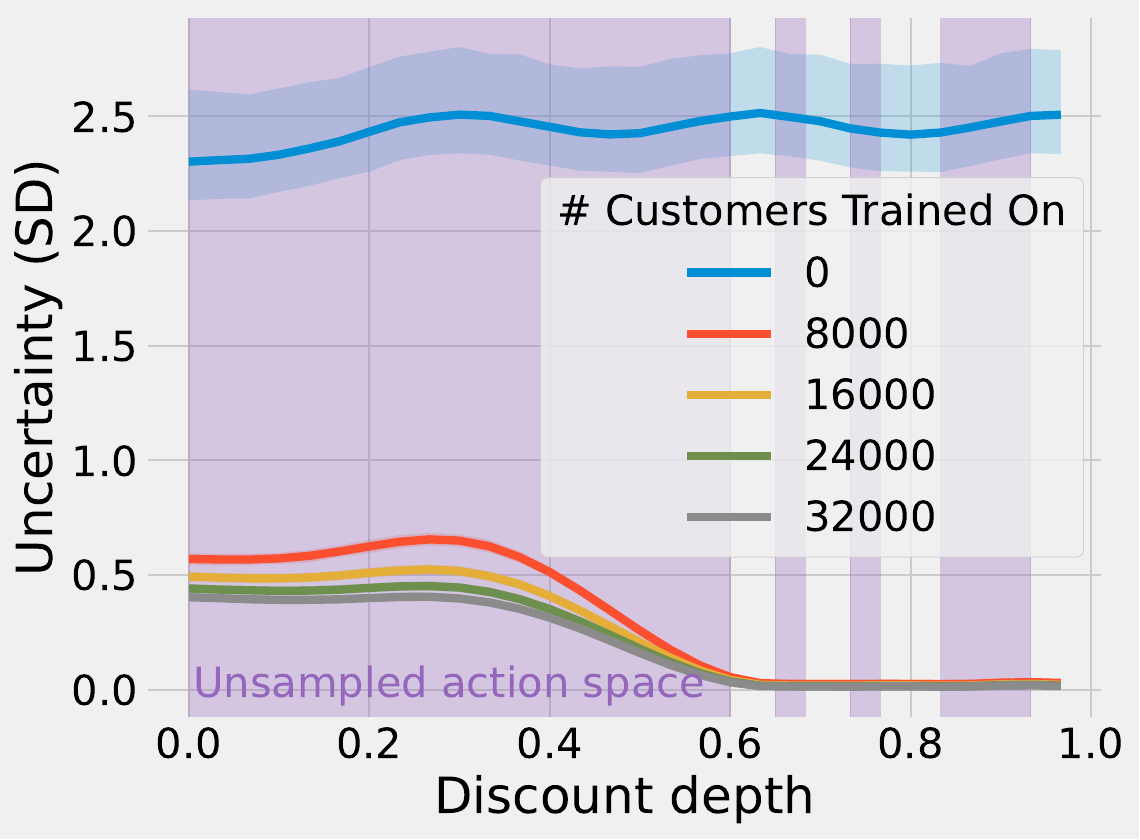} 
    % \caption{}
    \label{fig:rbf3}
\end{subfigure}
\caption{Action encoding mechanism.
The left figure illustrates a 3-dim encoding of each discount depth from 0.0 to 1.0 using the RBF transformation with three basis locations (0.25, 0.5, 0.75). This encoding mechanism leads to information sharing as measured by the effective number of times the algorithm has selected each action for a fixed context, depicted in the middle figure. The right figure demonstrates the uncertainty (standard deviation; SD) in the reward model adapts to increasing exposure to different regions of the action space, including regions that are unrepresented in the training data (extrapolation/interpolation; shaded in pink). Each line shows the uncertainty over 1K randomly selected customers, where the model is trained on different volumes of data. As the volume of data increases, the model retains greater uncertainty for the previously unseen extrapolation range $a < 0.6$. Meanwhile, its confidence still incrementally increases due to the RBF's information sharing.}
\label{figure: rbf_all}
\end{figure*}

We configured RBFs based on their ability to (a) support good predictive accuracy (measured by weighted absolute percentage error; WAPE), (b) capture the monotonicity between depth and full price basket values (Figure \ref{figure:monotonicity} (left); see Section \ref{sec:intro} for background). Figure \ref{figure: rbf_all} (left) shows the full action space as represented using 3 radial basis functions at [0.25, 0.50, 0.75], as well as the number of times each action was perceived by the algorithm for a fixed context when played 1K times at [0.40, 0.60, 0.80] (middle). This illustrates a major strength of the RBF encoding scheme: it allows the model to gain information about actions that are similar to those previously encountered, in order to generate future predictions.  

% \begin{figure*}[t]
%     \centering
% \begin{subfigure}[b]{0.32\textwidth}
%     \centering
%     \includegraphics[width=\textwidth]{figures/RBF1.pdf} 
%     \caption{}
    
%     \label{fig:rbf1}
% \end{subfigure}
% \hfill
% \begin{subfigure}[b]{0.32\textwidth}
%     \centering
%     \includegraphics[width=\textwidth]{figures/RBF2.pdf} 
%     \caption{}
%     \label{fig:rbf2}
% \end{subfigure}
% \hfill
% \begin{subfigure}[b]{0.32\textwidth}
%     \centering
%     \includegraphics[width=\textwidth]{figures/RBF3_uncertainty.pdf} 
%     \caption{}
%     \label{fig:rbf3}
% \end{subfigure}
% \caption{Action encoding. Encoding of the action space with three radial basis functions (RBFs; loc=0.25, 0.5, 0.75) (\ref{fig:rbf1}) leads to information sharing as measured by the effective number of times the algorithm has selected each action for a fixed context (\ref{fig:rbf2}). This enables information sharing between similar actions and a wider cover of the action space. (\ref{fig:rbf3}) Uncertainty (standard deviation; SD) in the reward model adapts to increasing exposure to different regions of the action space, including regions that are unrepresented in the training data (extrapolation/interpolation; shaded in pink). Each line shows the uncertainty over 1K randomly selected customers, where the model is trained on different volumes of data. As the volume of data increases, the model retains its greater uncertainty for the previously unseen extrapolation range $a < 0.6$, but does still incrementally increase its confidence due to the RBF's information sharing.}
% \label{figure: rbf_all}
% \end{figure*}

\subsection{Context feature representation}\label{section: feature mapping}

Although e-commerce businesses have access to many customer signals (e.g. historical spend, site interaction), 
directly adding them as features can harm the efficiency of learning due to the curse of dimensionality \cite{BanditsBook}.
% feature sets cannot be naively included as (interacting) context features without severely harming the efficiency of learning (i.e. due to the curse of dimensionality; \cite{BanditsBook}). 
To overcome this, we used a deep neural network (DNN) to predict the log full price basket value for each customer, and extracted lower dimensional representations from the penultimate layer for use in the downstream reward model (Figure \ref{figure:DISCO}). The DNN effectively serves as a function for representation learning $\psi_1: \mathcal{X}\rightarrow \mathbb{R}^{d_1}$, producing an abstract representation of its inputs.

The DNN was trained on ~5M customers who were active in a three-month period, using each customer's historical data over the preceding one year (including non-discounted purchases). N=76 features were fed into the model, including customer's purchase history (e.g. total/average spend), return history, discount code usage (e.g. average depth of used codes), and site interaction data (e.g. add-to-bag). The DNN consisted of four layers of sizes [64, 16, 6, 1]. The penultimate layer played a crucial role by extracting a 6-dimensional contextual embedding, effectively capturing the intricacies of the customer's purchasing patterns (performance was similar if dimensionality +/- 2) (Figure \ref{figure:DISCO}). DNN training was via mini-batch stochastic gradient descent, using the Adam optimizer (learning rate=0.001) and dropout regularization to reduce overfitting.

This feature representation $\mathcal{X}\rightarrow \mathbb{R}^{d_1}$ is a mapping of contextual features $\mathcal{X}$ and does not encompass the action space. It is worth noting that the extensive purchase data necessary for training the DNN can be obtained through normal operations, without requiring the retailer to run new discount campaigns. 

\subsection{Reward prediction: Bayesian log-linear regression}

After extracting context representations $\psi_1$ and action representations $\psi_2$, we build the final feature set for customer $i$ with discount $a$ by taking all possible pairwise products between $\psi_1$ and $\psi_2$:
\begin{equation}
\label{eq:regfullregression}
\small
    \psi\left(X_{t, i}, a\right) = \left\{\psi_1 \left(X_{t, i}\right), \psi_{2}\left(a\right), \psi_1 \left(X_{t, i}\right)  \times \psi_{2}\left(a\right)\right\} \in \mathbb{R}^{d}
\end{equation}
where $d = d_1 + d_2 + d_1 d_2$ and $\times$ denotes the Cartesian product between the two sets of feature mappings. Doing this gives us a rich class of policies where the optimal discount depth is dependent on the customer embedding vector. Using this feature set, we modelled log full price basket values as:
 \begin{equation}\label{eq:lm}
 \small
    \mathbb{E}\left[\ln\left(F_{t, i, a}\right)\,\vert \, X_{t, i} = x\,,\, A_{t, i} = a \right] = \langle \theta \,, \, \psi\left(x, a\right) \rangle
\end{equation}
We trained a Bayesian log-linear model using customer purchase data from periods that overlapped with discount code campaigns. Data from two campaigns were used for model training, with future campaigns used for testing. Training data was restricted to active customers who had made $\geq1$ purchase in the previous one year, and contained campaigns where the allocation of discount code to different customers had a large random component (see \cite{gruson19offlineeval} for an alternative methodology when using highly skewed historical datasets). The contextual feature embeddings were derived by applying the trained DNN to customer data from the week before the target campaign.

% For convenience, we refer to $Y_{t,i, a}$ to denote the reward quantity at time $t$ for customer $i$ under discount depth $a$, without specifying its exact formulation $h(.)$, except to note that it is distinct from (but also directly compute-able as a function of) the full price basket values $\tilde{F}_{t,i, a}$ output by the Bayesian log-linear regression model: 
% \begin{equation}
% \label{eq:abstractreward}
% \small
%     Y_{t,i, a} = h(F_{t,i, a}) \\
% \end{equation}

\subsubsection{Reward sampling}

We chose a linear reward model to enable \textbf{Thompson Sampling} (TS) \cite{ts_with_lin_payoffs}, which balances the explore/exploit trade-off by maintaining a posterior distribution over the parameter vector, $\theta$. The posterior distribution quantifies the model's uncertainty, and TS samples a pseudo-reward $\tilde{F}_{t, i, a}$ for each action $a \in \mathcal{A}_t$. Exploration is driven by uncertainty: as more information is acquired, the posterior distribution becomes more defined, leading to a reduced exploration. To facilitate computation of the inverse, we use the closed-form posterior with Gaussian priors over coefficients of the linear model \cite{ts_with_lin_payoffs}: 

% ------------------------- [V1] -------------------------
% \begin{equation}
% \small
%     \hat{\theta}_{t} \sim \mathcal{N}\left(\mu = \bar{V}_{t}^{-1} B_{t} \,, \sigma^2 = \beta_{t}^2 \bar{V}_{t}^{-1}\right)
%     % \ln\left(Y_{t, i}\right)\, \vert \psi\left(x, a\right) \stackrel{d}{=} \mathcal{N}\left(\mu = \langle \hat{\theta}, \psi\left(x, a\right)\rangle\,, \sigma^2 = \beta_{t}\norm{\psi\left(x, a\right) }_{\bar{V}_{t - 1}^{-1}}\right) 
% \end{equation}
% where  
% \small
% \begin{align*} 
%     \bar{V}_{t} &= V_{0} + \sum_{s = 1}^{t}\sum_{i = 1}^{I} \psi\left(X_{s, i}, A_{s, i}\right) \psi\left(X_{s, i}, A_{s, i}\right)^{T}\\
%     &= \bar{V}_{t - 1} + \sum_{i = 1}^{I} \psi\left(X_{t, i}, A_{t, i}\right) \psi\left(X_{t, i}, A_{t, i}\right)^{T}
% \end{align*}
% ------------------------- [V2] -------------------------
\begin{equation}
\small
    \hat{\theta}_{t} \sim \mathcal{N}\left(\mu = \bar{V}_{t}^{-1} B_{t} \,, \sigma^2 = \beta_{t}^2 \bar{V}_{t}^{-1}\right)
    % \ln\left(Y_{t, i}\right)\, \vert \psi\left(x, a\right) \stackrel{d}{=} \mathcal{N}\left(\mu = \langle \hat{\theta}, \psi\left(x, a\right)\rangle\,, \sigma^2 = \beta_{t}\norm{\psi\left(x, a\right) }_{\bar{V}_{t - 1}^{-1}}\right) 
\end{equation}
where,
\small
\begin{equation} 
\begin{split}
    \bar{V}_{t} &= V_{0} + \sum_{s = 1}^{t}\sum_{i = 1}^{I} \psi\left(X_{s, i}, A_{s, i}\right) \psi\left(X_{s, i}, A_{s, i}\right)^{T}\\
    & = \bar{V}_{t - 1} + \sum_{i = 1}^{I} \psi\left(X_{t, i}, A_{t, i}\right) \psi\left(X_{t, i}, A_{t, i}\right)^{T}\\
\end{split}
\label{eq:V_t}
\end{equation}
with $V_0^{-1}$ (typically the identity matrix) being the prior precision matrix, $\beta_t$ being the exploration hyperparameter, and
\small
\begin{equation} 
\begin{split}
    B_t & = \sum_{s = 1}^{t}\sum_{i = 1}^{I} \psi\left(X_{s, i}, A_{s, i}\right) \ln(F_{s,i,a})\\
    & = B_{t-1} + \sum_{i = 1}^{I} \psi\left(X_{t, i}, A_{t, i}\right) \ln(F_{t,i,a})\\
\end{split}
\label{eq:B_t}
\end{equation}

Thus, we can efficiently maintain the posterior distribution using the Woodbury matrix identity, which requires $\mathcal{O}(d^2)$ operations and $\mathcal{O}(d^2 + d)$ space (an improvement over MCMC). For each customer, we sample $\tilde{F}_{t, i, a}$ for all $a\in\mathcal{A}_t$ from the posterior and apply the exponent to return appropriate units. This sampling strategy prevents the learner from consistently selecting the greedy action, and ensures sufficient exploration in each round. Each batch of $\tilde{F}_{t, i, a} \forall a\in\mathcal{A}_t$ in round $t$ is then fed to the downstream constrained integer program, for decision making. 

% \begin{table}
%   \caption{Sampled pseudo-reward for each customer (row) and action (column) for a fixed round $t$, computed using Thompson Sampling with the Bayesian log-linear regression. Pseudo-rewards are sampled from the posterior distribution to populate values in a lookup table, which is then fed into a downstream integer program for discount code allocation.
%   }
%   \label{table: thompson sampling output}
%   \small
%   \begin{tabular}{ccccc}
%     \toprule
%     Customer& $a_1$ & $a_{2}$ & $\cdots$ & $a_{K}$\\
%     \midrule
%     $1$ & $\tilde{Y}_{1, a_1}$ & $\tilde{Y}_{1, a_2}$  & $\cdots$ & $\tilde{Y}_{1, a_K}$ \\
%     $2$ & $\tilde{Y}_{2, a_1}$ & $\tilde{Y}_{2, a_2}$  & $\cdots$ & $\tilde{Y}_{2, a_K}$ \\
%     $\vdots$ & $\vdots$ & $\vdots$ & $\cdots$ & $\vdots$\\
%     $I$ & $\tilde{Y}_{I, a_1}$ & $\tilde{Y}_{I, a_2}$  & $\cdots$ & $\tilde{Y}_{I, a_K}$ \\
%   \bottomrule
% \end{tabular}
% \end{table}

\subsection{Optimisation of discount code allocation}

Adhering to the traditional application of Thompson Sampling involves selecting the action that yields the highest reward per customer. However, this would ignore important business constraints, such as the markdown budget, or the need to control the \textit{range} of experiences offered to customers. This latter concern is common in customer-facing retail contexts, where businesses need to manage their brand and customer relationships by taking a holistic view. To allow for such holistic control, we formulate discount code allocation as an integer program that takes the target discount depth \textit{distribution} in as an input constraint from the operational team. 

For a discount campaign $t$ with the discount depths $\mathcal{A}_t$ and $\tilde{F}_{i, a}$ for each customer-action combination, discounts were allocated via the following integer program:
\begin{equation}
\label{eq:LP}
\small
\begin{array}{ll@{}ll}
\text{Maximise}  &  \displaystyle\sum\limits_{i = 1}^{I} \sum\limits_{a\in \mathcal{A}_t} (w \cdot \tilde{R}_{i, a} - \tilde{C}_{i, a})  \cdot {s_{i, a}} \cdot e_{a}& \\
\small
\text{subject to:}   & {s_{i,a}} \in \{0, 1\} \hspace{1cm} \forall (i, a) \in \{1, 2, \cdots, I\} \times \mathcal{A}_t\\
                    & \displaystyle\sum\limits_{a\in \mathcal{A}_t} {s_{i,a}} \le 1 \hspace{1cm} \forall i \in \{1, 2, \cdots I\}\\ 
                    & \displaystyle\sum\limits_{i = 1}^{I} {s_{i,a}} \le N_{a} \hspace{1cm} \forall a \in \mathcal{A}_t\\
\end{array}
\end{equation}

\noindent where $\tilde{R}_{i, a}$ is the expected revenue for customer $i$ offered discount $a$ (calculated $R_{t,i,a} = F_{t,i,a} * (1-A_{t,i})$), $\tilde{C}_{i, a}$ is the expected markdown cost (calculated $C_{t,i,a} = F_{t,i,a} * A_{t,i}$), $w$ is an importance weight used by operators to control the priority of revenue-maximisation (vs cost minimization) goals in the campaign, $s_{i, a}$ is a binary variable indicating whether a customer $i$ is offered the discount depth $a$, $N_{a}$ is the number of users allocated to $a\in\mathcal{A}_t$, and $e_a$ is the engagement rate of discount depth $a$ (proportion of customers who completed purchases with the allocated code, out of the number of customers who received it; historical averages were used to compute $e_a$). The distribution of $N_{a}$ is specified by stakeholders for each $a\in\mathcal{A}_t$ in every round to control the overall distribution of discount depths. Note that Eq.\ref{eq:LP} allows one to tactically adjust the relative priority of maximising revenue versus reducing cost, by changing the $w$ parameter for each campaign. Additionally, although DISCO allows operators to specify the distribution over different depths, Eq.\ref{eq:LP} can be easily adapted to incorporate the budget as an additional constraint, providing further flexibility.

\section{Experiments}
\label{sec:experiments}

To assess the performance of DISCO, we performed offline analyses focusing on different aspects of the algorithm. \textbf{For commercial sensitivity, all discount depths, revenue, basket value numbers, and \% increase in basket values reported have been rescaled to arbitrary units}.

\subsection{Information sharing and price elasticity with RBF encoding}
\label{sec:monotonic}

% \begin{figure}
%     \centering
%     \includegraphics[scale = 0.20]{figures/monotonicity_all.png}
%     \caption{Negative price elasticity. (A) shows the observed relationship between discounting and \textit{full price} basket values, which is in line with the conventional assumption of price elasticity. Monotonicity is expected and observed only when looking at full price basket values, not discounted ones. (B) Different action encoding mechanisms and their effects. An RBF encoding scheme with $K=3$ centroids and $\alpha=20$ demonstrates the desired near monotonic relationship between the actions and their corresponding effects. (C) The chosen action encoding scheme (K=3, $\alpha=20$) produced the expected monotonicity as used in the overall Bayesian log-linear reward model, both overall (blue; CIs indicate 95\% CI of the mean) as well as for 3 randomly selected customers.
%     }
%     \label{figure:monotonicity}
% \end{figure}

\begin{figure*}[t]
    \centering
\begin{subfigure}[b]{0.32\textwidth}
    \centering
    \includegraphics[width=\textwidth]{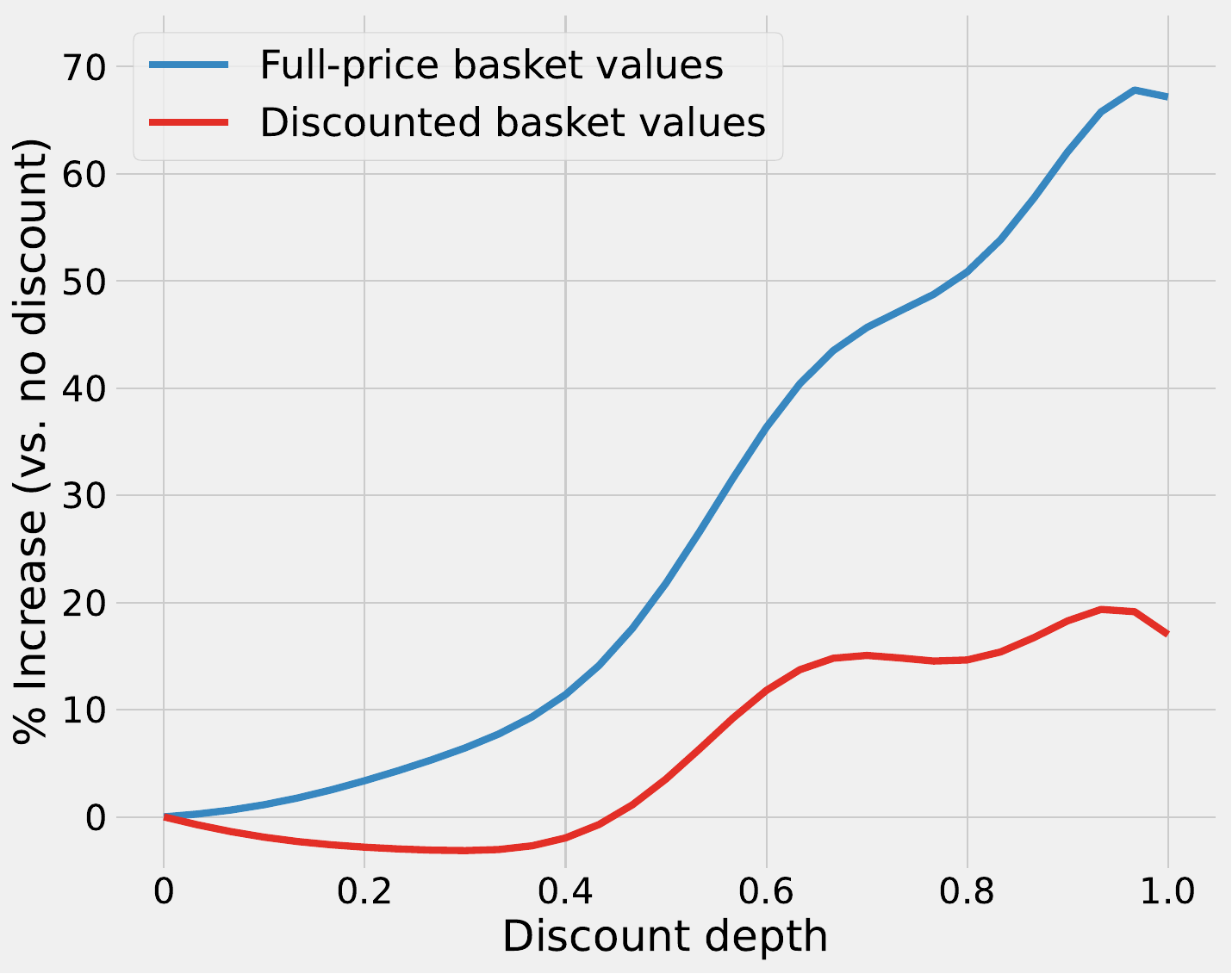} 
    % \caption{}
    \label{fig:price_elasticity1}
\end{subfigure}
\hfill
\begin{subfigure}[b]{0.32\textwidth}
    \centering
    \includegraphics[width=\textwidth]{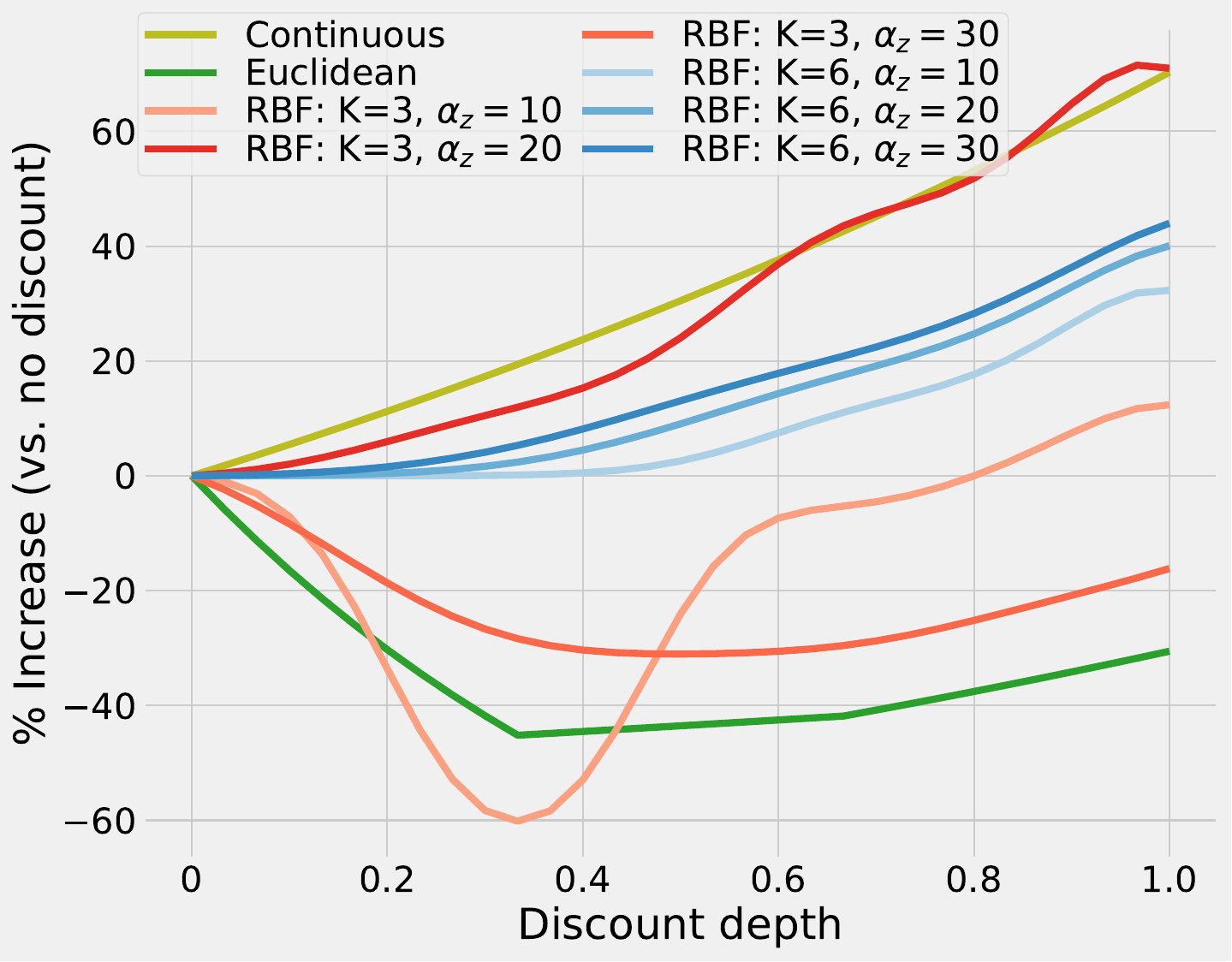} 
    % \caption{}
    \label{fig:price_elasticity2}
\end{subfigure}
\hfill
\begin{subfigure}[b]{0.32\textwidth}
    \centering
    \includegraphics[width=\textwidth]{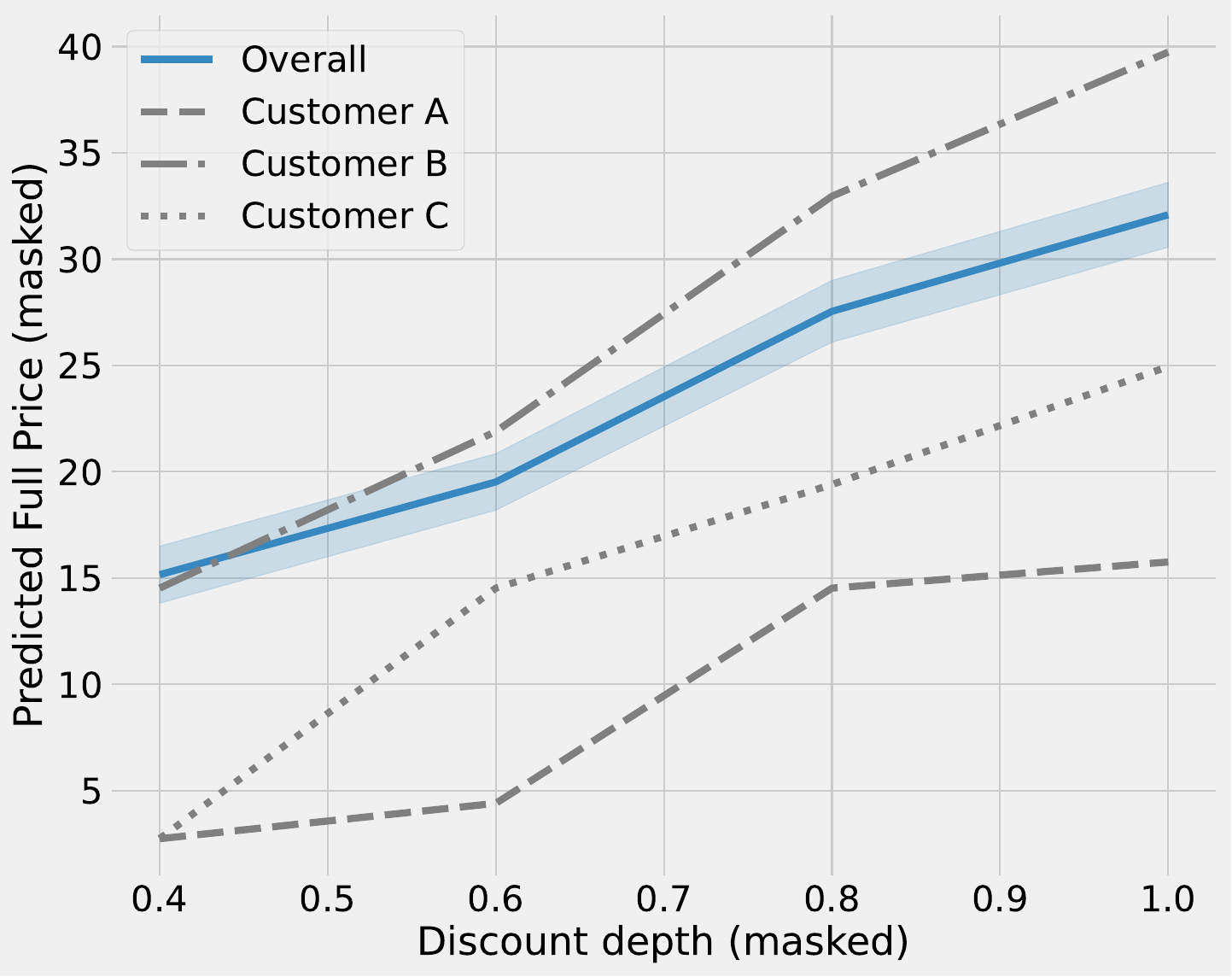} 
    % \caption{}
    \label{fig:price_elasticity3}
\end{subfigure}
\caption{Negative price elasticity. The left figure shows the observed
relationship between discounting and full-price basket values, which is in line with the conventional assumption of
price elasticity. Monotonicity is expected and observed only
when looking at full-price basket values, not discounted ones. The middle figure demonstrates different action encoding mechanisms and their effects. An RBF encoding scheme with $K=3$ centroids and $\alpha=20$ demonstrates the desired near-monotonic relationship between the actions and their corresponding effects. On the right figure, the chosen action encoding scheme (K=3, $\alpha=20$) produced the expected monotonicity as used in the overall Bayesian log-linear reward model, both overall (blue; CIs indicate 95\% CI of the mean) as well as for 3 randomly selected customers.}
\label{figure:monotonicity}
\end{figure*}

\textbf{Accuracy and negative price elasticity.} Figure \ref{figure:monotonicity} (left) shows the relationship between discount depth and full price basket values, as observed in our own dataset (as well as in line with conventional assumptions around negative price elasticity) (see Section \ref{sec:intro}). As mentioned in Section \ref{sec:problem}, we focused our modelling on preserving the expected monotonicity and price elasticity with respect to discount depths and \textit{full price} basket values. To configure the RBFs in the reward prediction model, we evaluated several different encoding schemes for the continuous action space, focusing on predictive accuracy, monotonicity (negative price elasticity), and low dimensionality. The various action encoding schemes considered provided similar performance in terms of accuracy (all WAPEs=0.140 at 3d.p. precision, Spearman's $\rho$=0.475 at 3d.p. precision). Figure \ref{figure:monotonicity} (middle) displays the different RBF and alternative encoding schemes considered and illustrates the uneven ability of different options to preserve the monotonicity (negative price elasticity) between actions and their corresponding effects. The first line represents continuous encoding, where actions are represented on a continuous scale. However, this encoding method is inadequate in capturing the inherent non-linear relationship between actions and effects. The second line represents Euclidean encoding, which measures the Euclidean distance between actions and a reference point. The next six lines depict the RBF encoding with varying numbers of centroids and $\alpha$ values. Notably, the red line and three blue lines using RBF encoding exhibit a desirable trend, closely approximating monotonicity between actions and their corresponding effects. Based on these observations, we employed RBFs with three centroids and $\alpha=20$ in our model. Although the seven-centroid options also exhibited monotonicity, the three-centroid configuration was preferable due to the lower dimensionality of the final feature set in the reward prediction model. 

Figure \ref{figure:monotonicity} (right) shows the expected negative price elasticity in the model's predictions for 1K randomly selected customers, as well as for three randomly-selected individual customers. While price elasticity can vary significantly both \textit{across} customers as well as across discount depths \textit{within} individual customers, the reward model was well able to preserve the assumption of price elasticity both in the general case as well as across the vast majority of state-action space (>90\%).

% Nevertheless, the reward model continued to output (out-of-sample) predictions in this $a < 0.6$ range that were both reasonable (see Figure \ref{figure:full_price}) as well as accurate (Table \ref{table:lm}). This is primarily due to the use of radial basis functions in action encoding, which allow for information sharing. 

% \begin{figure}
%     \centering
%     \includegraphics[scale = 0.4]{figures/uncertainty.pdf}
%     \caption{
%     Uncertainty (as measured by standard deviation; SD) in the reward model adapts to increasing exposure to different regions of the action space. Each line shows the uncertainty of 1,000 randomly selected customers, where the model is trained on different volumes of data. Depths of $a < 0.6$ were absent from the training dataset. As the volume of data increases, the model retains its greater uncertainty for the $a < 0.6$ range (vs. $a>0.6$) but does still incrementally increase its confidence within this range, due to information sharing enabled by the RBF encoding.
%     % For commercial sensitivity,
%     % all discount depths reported are scaled to arbitrary units.
%     }
%     \label{figure:uncertainty}
% \end{figure}

\textbf{Uncertainty.} The use of RBFs enabled information sharing and efficient non-linear learning, resulting in highly accurate predictions, including for action values that had not been observed in the historical data (see Section \ref{results:rewardpredmodel}). 
We were also interested in how model uncertainty was attenuated as the model was exposed to more data. Figure \ref{figure: rbf_all} (middle) compares uncertainty expressed by the Bayesian log-linear model (as measured by standard deviation, SD; calculated by sampling predicted basket value 1K times) for 1K randomly selected customers, where the models were trained on different amounts of historical data. It is worth noting that all batches of training data exclusively consisted of depths greater than 0.6 ($a > 0.6$). Consequently, the uncertainty estimates shown in Figure \ref{figure: rbf_all} (middle) for depths lower than 0.6 ($a < 0.6$) reflect the model's uncertainty in extrapolation. Despite this extrapolation, the model maintained a high level of predictive accuracy, aided by the RBFs (see Section \ref{results:rewardpredmodel}). Additionally, we observed that the reward prediction model appropriately attenuated its confidence as it gained exposure and displayed increased uncertainty for depths it had not encountered in the training data (i.e. $a < 0.6$, compared to $a > 0.6$). This relatively higher uncertainty in extrapolation is both anticipated and advantageous, given the model's lack of exposure to the $a < 0.6$ range. More broadly, the model trained with larger datasets exhibits reduced uncertainty in its predictions, as expected.

\subsection{Reward prediction model}
\label{results:rewardpredmodel}

\textbf{Contextual representation with DNN.}
The DNN was evaluated on customer purchases in one calendar month after the training period. We contrast DNN performance against three other popular regression models: Least Square Regression (LR), Light GBM (LGBM), and Random Forest (RF). DNN (WAPE=0.153, Spearman Correlation $\rho$=0.409) demonstrated similar accuracy in predicting \textit{full price} basket values compared to RF (WAPE=0.153, $\rho$=0.409) and LGBM (WAPE=0.154, $\rho$=0.410) while outperforming LR (WAPE=0.160, $\rho$=0.346) significantly. Despite the comparable accuracy of the RF and LGBM models, the DNN is more suitable for the primary objective of generating context embeddings for downstream systems.

% \begin{table}
%   \caption{Performance of the Bayesian log-linear regression model in predicting (log) full price basket values across four different discount code campaigns. The model was trained on data from Campaigns A and B, and its robust performance in Campaigns C and D demonstrates successful generalisation. In particular, the model maintained high accuracy in Campaign D, which is only comprised by unseen depths.
%   }
%   \label{table:lm}
%   \small
%   \begin{tabular}{ccccc}
%     \toprule
%      & Campaign A& Campaign B& Campaign C & Campaign D\\
%     % Customer& $a_1$ & $a_{2}$ & $\cdots$ & $a_{K}$\\
%     \midrule
%     Wape & 0.140& 0.140& 0.139& 0.134\\
%     Spearman &  0.476& 0.447& 0.438& 0.461\\
%     % $1$ & $\tilde{Y}_{t, 1}^{a_1}$ & $\tilde{Y}_{t, 1}^{a_2}$  & $\cdots$ & $\tilde{Y}_{t, 1}^{a_K}$ \\
%     % $2$ & $\tilde{Y}_{t, 2}^{a_1}$ & $\tilde{Y}_{t, 2}^{a_2}$  & $\cdots$ & $\tilde{Y}_{t, 2}^{a_K}$ \\
%     % $\vdots$ & $\vdots$ & $\vdots$ & $\cdots$ & $\vdots$\\
%     % $B$ & $\tilde{Y}_{t, B}^{a_1}$ & $\tilde{Y}_{t, B}^{a_2}$  & $\cdots$ & $\tilde{Y}_{t, B}^{a_K}$ \\
%   \bottomrule
% \end{tabular}
% \end{table}

\textbf{Reward prediction model (Bayesian log-linear regression).} The final reward prediction model (Bayesian log-linear regression) was trained solely on two random campaigns, and showed high accuracy when tested on a new unseen campaign of a similar type (WAPE=0.139, Spearman's $\rho=0.438$). We were also interested in the model's performance when applied to a completely different type of campaign that differed in customer approach (email vs on-site), code redemption time (single use with a month-long redemption window, as opposed to the typical 1-2 days), as well as in the discount depths that were offered. This effectively tested the model's ability to generalize, both in terms of its ability to capture a customer's consistent behaviour across different touchpoints, as well as to new actions: this new campaign specifically consisted of depths that were \textit{shallower} than the depths observed in the training data, and therefore required the model to extrapolate (rather than interpolate) beyond the actions that it had previously observed in the training data. Despite these differences, the model maintained its good performance (WAPE=0.134, Spearman's $\rho=0.461$). This indicates that our models successfully captured the underlying relationship between depth and subsequent purchases, enabling accurate generalisation and extrapolation (albeit with higher uncertainty; see Section \ref{sec:monotonic}). Overall, DISCO's model is able to (1) identify and rank big and small spenders correctly (as indicated by a Spearman rank correlation) and (2) predict customer revenue accurately across different types of discount campaigns with previously unseen depths.

\subsection{Active learning with global constraints}

% \begin{figure}
%     \center
%     \includegraphics[width=\textwidth]{figures/bandit_simus_3.png} 
%     \caption{Evaluation of bandit algorithms. (A) Performance of different \textit{constrained} agents under warm- (left) and cold-start (right) scenarios. TS-IP demonstrates the strongest long-term performance, while UCB-IP's long term performance is notably hampered. (B) Comparison of TS-IP to a TS-ULCC benchmark ("Unconstrained Learner, Constrained Consumer"; warm start) whose "exploitative" actions are IP-constrained, but who takes separate "explorative" actions and observes rewards to update the model, without consuming rewards as reported above. Consumed rewards reported above stem from IP-constrained actions, using the predictive model that has benefited from unconstrained-action updates over time. Benchmarking against TS-ULCC quantifies the extent to which TS-IP's long-term performance is hampered by not being able to choose actions over the full action space (due to the IP constraint), while controlling for the reality of pragmatic action constraints as it relates to harvested rewards in each round. While TS-IP's long-term performance is indeed degraded relative to ULCC's (idealized) benchmark, the extent of degradation is very small (0.234\%), and does not dramatically escalate over 100 rounds of learning. This indicates that our IP constraint is not having an unacceptably deleterious effect on DISCO's ability to enact active learning.
%     }
%     \label{figure:bandit_sims}
% \end{figure}

\begin{figure*}[t]
    \centering
\begin{subfigure}[b]{0.32\textwidth}
    \centering
    \includegraphics[width=\textwidth]{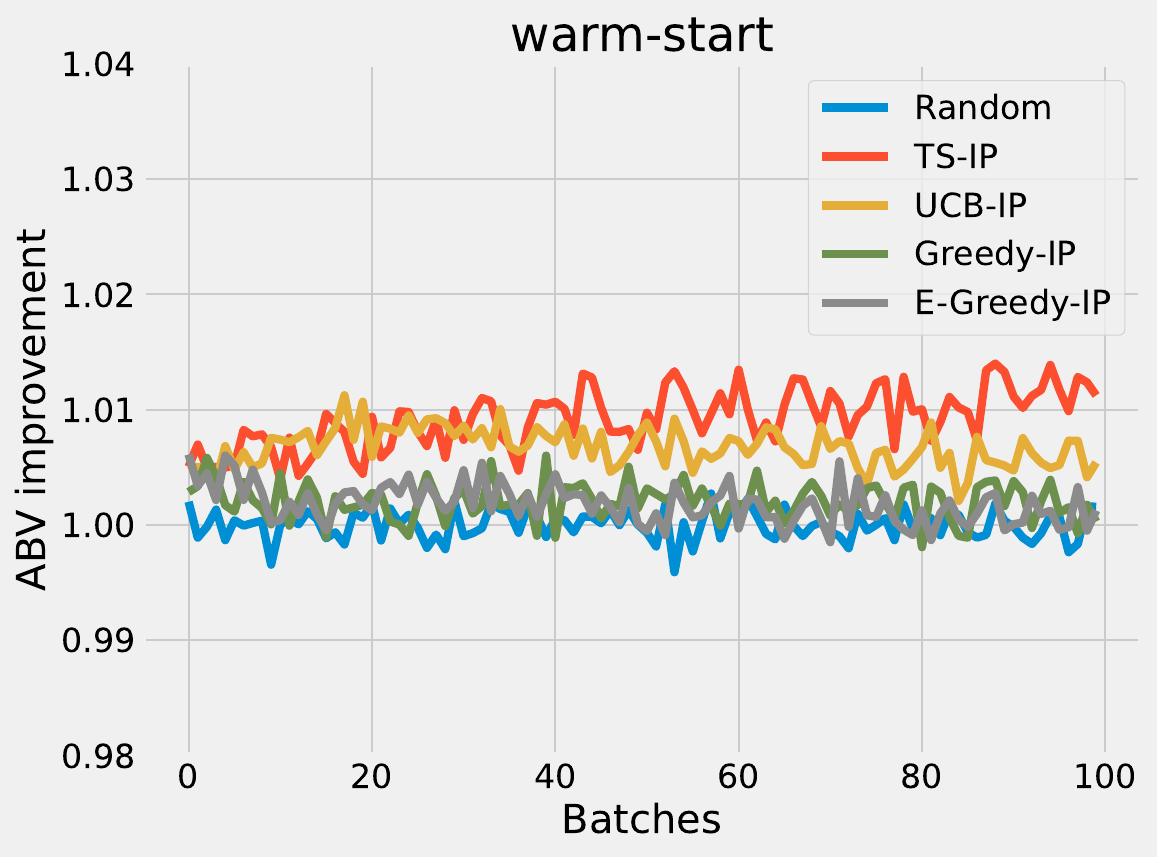} 
    % \caption{}
    \label{fig:bandit_sims1}
\end{subfigure}
\hfill
\begin{subfigure}[b]{0.32\textwidth}
    \centering
    \includegraphics[width=\textwidth]{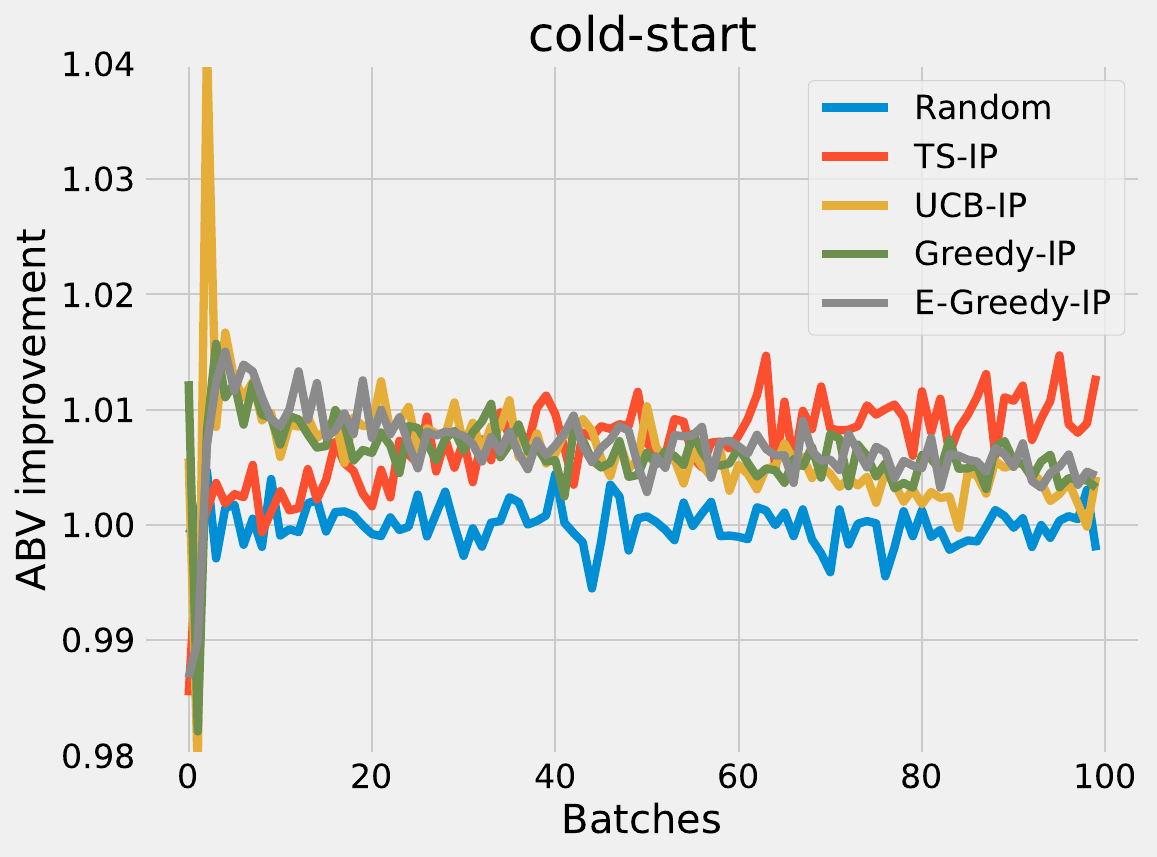} 
    % \caption{}
    \label{fig:bandit_sims2}
\end{subfigure}
\hfill
\begin{subfigure}[b]{0.32\textwidth}
    \centering
    \includegraphics[width=\textwidth]{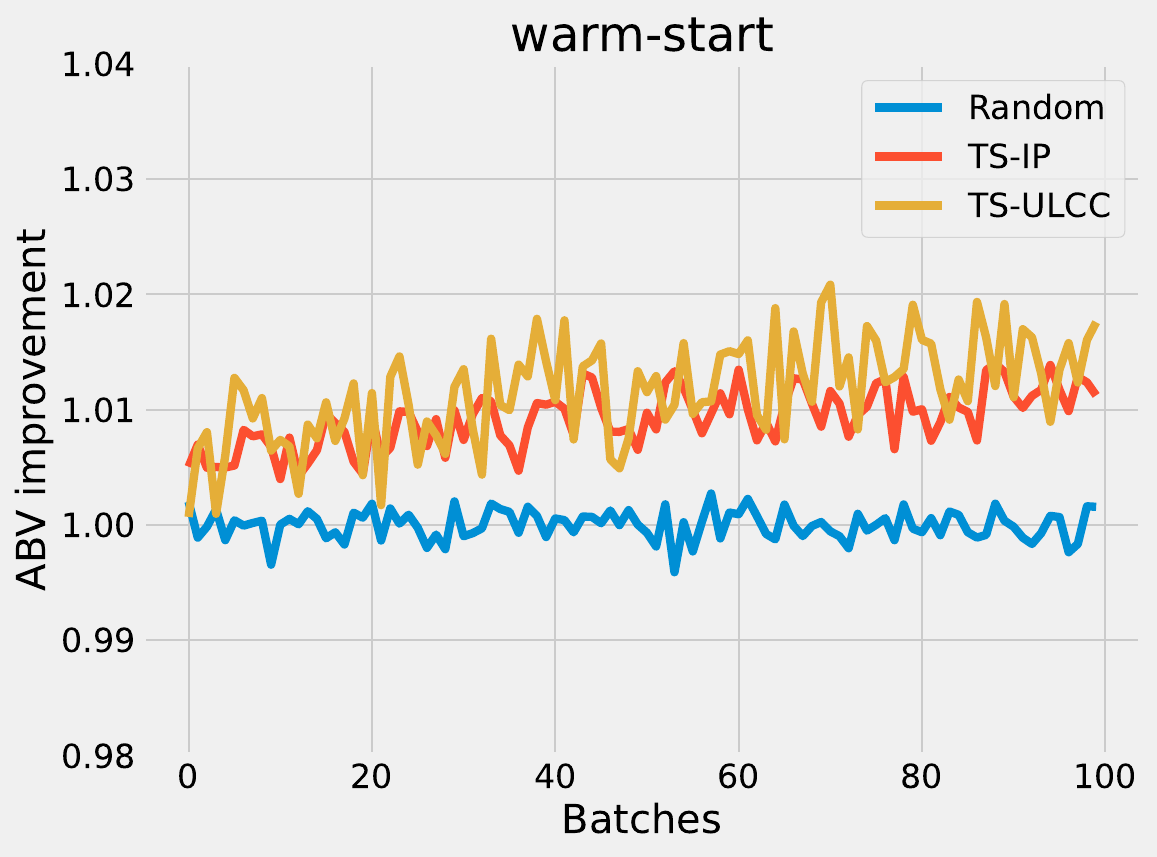} 
    % \caption{}
    \label{fig:bandit_sims3}
\end{subfigure}
\caption{Evaluation of bandit algorithms. Performance of different \textit{constrained} agents under warm- (left) and cold-start (middle) scenarios. TS-IP demonstrates the strongest long-term performance, while UCB-IP's long term performance is notably hampered. The right figure compares TS-IP to a TS-ULCC benchmark ("Unconstrained Learner, Constrained Consumer"; warm start). In this benchmark, "exploitative" actions are IP-constrained, but separate "explorative" actions are taken to update the model without consuming rewards. The consumed rewards reported earlier come from IP-constrained actions, using a predictive model enhanced by unconstrained-action updates over time. Benchmarking against TS-ULCC quantifies how much TS-IP’s long-term performance is affected by the inability to choose actions across the full action space (due to the IP constraint), while considering practical action constraints related to harvested rewards in each round. Although TS-IP’s long-term performance is slightly degraded compared to ULCC's idealized benchmark, the degradation is minimal (0.234\%) and does not significantly escalate over 100 rounds of learning. This indicates that the IP constraint does not have an unacceptably harmful effect on DISCO’s active learning capabilities.}
\label{figure:bandit_sims}
\end{figure*}

DISCO differs from traditional active learning in that actions are subjected to global constraints, which is a very common requirement of practical applications. In our experiments, we sought to assess how bandit algorithms would perform when subject to such constraints. Additionally, we were interested in evaluating the extent to which our bandit algorithm's learning ability might be degraded by the constraining of choice that stemmed from the integer program. Because initial analysis with theoretical environments indicated that results were highly sensitive to the configuration of the agent's environment, we quickly re-focused our efforts towards studying algorithms under \textit{realistic} distributions of consumer behaviour, by using real data from a genuine campaign. We adopted a standard process for producing unbiased offline, off-policy estimates of algorithm performance (\cite{li2010contextual}): using a genuine campaign in which discounts were \textit{randomly} assigned across customers, we (a) re-sampled customers to create a dataset in which discounts were uniformly distributed (in addition to being randomly assigned), (b) used rejection sampling to estimate rewards that each different algorithm would be expected to achieve under the action-constrained situation (see \cite{li2010contextual} for detail).  

In addition to evaluating a constrained variant of the \textbf{Thompson Sampling} (TS-IP) algorithm, we also evaluated constrained versions of other popular algorithms including \textbf{Upper Confidence Bound } (UCB-IP) \cite{Abbasi-Yadkori2011}, $\epsilon$-\textbf{Greedy} (E-Greedy-IP), a \textbf{Greedy} baseline (Greedy-IP), and a \textbf{Random} baseline. All algorithms (except for Random) were constrained by the integer program (Eq. \ref{eq:LP}), using realistic parameters obtained from a recent campaign. To assess performance, we looked at average basket value (ABV), which considers discounted value and reflects the revenue received by the company. The model was updated in sequential batches of 5K customers, and the results of the offline simulations were based on the average of 100 iterations of the Monte Carlo process for each algorithm. In the cold-start scenario, the algorithms had no prior information about customer behaviour. In the warm-start scenario, the algorithms were trained using a separate dataset from an earlier campaign.

Figure \ref{figure:bandit_sims} shows the efficacy of each algorithm under both warm-start (left) and cold-start (middle) scenarios, with all results scaled relative to the random policy (with the average of the random policy set to 1). In both scenarios, TS-IP demonstrated successful learning and improvement of ABV over time. TS-IP also outperformed greedy policies in both scenarios, although this required several learning batches to achieve in cold start. While the Greedy and $\epsilon$-Greedy approaches initially showed good performance (relative to Random) after an initial warm start, both of these algorithms declined in performance over time, likely due to the biased datasets collected by the Greedy-IP, which ultimately skewed model's performance. This clearly demonstrates the advantages of using active learning approaches over greedy ones, even though the latter may exhibit initial benefits. Interestingly, the trajectory of TS-IP indicated consistent improvement over the course of 100 batches in both the cold- and warm-start scenarios. In contrast, UCB-IP performance declined over time, indicating that UCB's exploration capabilities were more severely impacted by the global IP constraint.

We were also interested in how TS-IP's ability to actively learn might be hampered relative to \textit{unconstrained} Thompson Sampling. While it would not have been meaningful to directly compare TS-IP to vanilla, unconstrained TS (because TS would be able to offer deeper discounts that would naturally lead to greater full price basket values), we sought to compare TS-IP to a different agent whose predictive model was updated by unconstrained TS actions (and knowledge of the subsequent rewards), but who then chose actions that were subject to the usual IP constraints. This algorithm, while artificial in that an agent would never be able to take different sets of actions in order to separately explore (learn) vs exploit, does give us a useful comparison point in which reward harvesting (exploit) is constrained, while active learning (exploration and model updating) is not. This comparison allows us to quantify the extent to which the benefits of active learning are degraded, as a result of TS-IP's constraints. Although the IP reflects genuine business considerations that cannot be entirely ignored, such benchmarking remains a useful exercise, as it can be used to assess the benefits of reconfiguring the constraints (e.g. by changing $N_a \forall \mathcal{A}$ in eq \ref{eq:LP}), or by re-formulating the constraints entirely) in order to better manage the explore-exploit trade-off. It is also more meaningful to compare TS-IP to this proposed agent rather than pure unconstrained TS, since the latter's unconstrained actions would always produce greater rewards by dint of its ability to take more aggressive actions - even in static, full information contexts for which no active learning needs to ever occur. 

To create this benchmark, we designed an algorithm that was able to make and learn from unconstrained actions, but whose \textit{consumptive} rewards came from actions that conformed to the constraints ("Unconstrained Learner, Constrained Consumer", ULCC). This algorithm consists of two components: (a) a "Learner" who takes unconstrained actions, and observes subsequent rewards that are only used to update the predictive model (b) a "Consumer", who takes actions that are subject to constraint, but with the ability to use the predictive model that has been iteratively updated by the Learner. Importantly, the rewards harvested by this algorithm are related to the (constrained) actions taken by the Consumer, reflecting a realistic relationship between the business constraints and the subsequent reward within each batch (even when there is perfect knowledge). Figure \ref{figure:bandit_sims} (right) shows the performance of TS-IP, relative to this idealized TS-ULCC algorithm comparison (warm start). Although TS-ULCC outperforms TS-IP as expected, the extent of the difference is very small: the overall drop in reward across 100 batches is 0.234\%, comparing TS-IP's ABV to TS-ULCC (ttest comparing rewards across TS-IP vs TS-ULCC, p<0.001). Additionally, this degradation did not appear to grow dramatically over time: ttests comparing rewards in the first vs last 50 batches of the simulation did not find a significant difference in the size of the reward degradation comparing TS-IP to TS-ULCC (p>0.05). Although in theory one might seek to capitalize on all performance improvements that are possible, such a small performance degradation (<1\%) is in practice a small price to pay for the benefits of operational control that are provided by the IP constraint, without which such a system would not be useable at all.

We have here introduced a method for benchmarking constrained bandit algorithms against their unconstrained versions, in order to evaluate a constraint's negative impact on active learning over time. These results demonstrate the ability of TS-IP to learn with reasonably good efficiency over an extended period of time, and we adopted this learning algorithm within DISCO. These results also highlight the importance of active learning (compared to greedy approaches) in achieving effective discount allocation over the long-term horizon, and highlight the superiority of TS over UCB bandit algorithms specifically in situations where global constraints apply. 

\section{Online A/B Test}

Finally, we tested the efficacy of the system by conducting a large-scale online A/B test during a discount code campaign at ASOS.com. In the campaign, all eligible customers were randomly assigned to the Test or the Control group, with the Test group's discounts determined by DISCO, and the Control group's discounts allocated randomly across customers but with the same cost control configuration (i.e. $N_{a}$ values in eq. \ref{eq:LP}). The Control group experienced an operational approach that is used in existing campaigns, that reduces campaign costs (relative to undifferentiated campaigns where all customers get the same discount) by controlling the distribution of discounts ($N_{a}  \forall  \mathcal{A}$), but without further optimization. Due to commercial sensitivity, we omit reporting of group averages and other aspects of customer behaviour, and instead focus on relative improvements: DISCO outperformed the Control in generating revenue by +1.12\% (p<0.001), and generated more reward by +1.23\% (p<0.01; reward=revenue-cost as shown in eq. \ref{eq:LP}). Additionally, DISCO's models maintained similar predictive accuracy in the online test as seen during offline evaluation (WAPE=0.133, Spearman's $\rho$=0.446), which indicates the veracity of the offline evaluation methods as well as of the models themselves. We note additionally that the extent of improvement shown here is roughly in line with what one might expect when observing the (warm start) offline simulations in Figure \ref{figure:bandit_sims} (left).

We are also able to measure DISCO's performance relative to (more commonly-used) undifferentiated discount campaigns in which all customers receive the same discount. Unconfounded measurement here is possible because customers experiencing this undifferentiated discount value effectively constitute a (randomly assigned, and thus unconfounded) subset of customers within the control condition alone. In this comparison, we find that DISCO outperforms the legacy undifferentiated discounts in both revenue (+3.56\%, p<0.001) as well as reward (+4.10\%, p<0.001). These results illustrate the importance of personalising discounts in optimizing operations, and demonstrate the efficacy of DISCO as a method for doing so.

% \begin{table} 
%     \caption{
%       Online A/B test. DISCO outperforms the control condition, as well as the legacy approach of distributing undifferentiated discounts to the entire customer base. For commercial sensitivity, percentage uplifts are shown rather than group averages. p-values are obtained via t-tests, with Welch's t-test used for the Legacy comparison (due to unequal variances). Online model performance matched offline model metrics (see \textcolor{red}{Table \ref{table:lm}}), indicating the accuracy of the predictive models overall. \textcolor{red}{TODO: for space, cam embed these into the text}
%       }
%   \label{table:ab} 
%   \small
%   \begin{tabular}{c|cc|cc}
%     \toprule
%      & Revenue & Reward (revenue-cost) \\
%     \midrule
%      vs. Control  &+1.12\%  (p<0.001) & 1.23\%  (p<0.01) \\
%      vs. Legacy (Undifferentiated) &+3.56\%  (p < 0.001) & 4.10\% (p<0.001) \\
%   \bottomrule
%  \end{tabular}
 
%  \begin{tabular}{ccccc}
%     \toprule
%     WAPE& Spearman& MAE & MSE & $R^2$\\
%     % Customer& $a_1$ & $a_{2}$ & $\cdots$ & $a_{K}$\\
%     \midrule
%     0.133& 0.446& 0.557& 0.496& 0.113\\
%     % $1$ & $\tilde{Y}_{t, 1}^{a_1}$ & $\tilde{Y}_{t, 1}^{a_2}$  & $\cdots$ & $\tilde{Y}_{t, 1}^{a_K}$ \\
%     % $2$ & $\tilde{Y}_{t, 2}^{a_1}$ & $\tilde{Y}_{t, 2}^{a_2}$  & $\cdots$ & $\tilde{Y}_{t, 2}^{a_K}$ \\
%     % $\vdots$ & $\vdots$ & $\vdots$ & $\cdots$ & $\vdots$\\
%     % $B$ & $\tilde{Y}_{t, B}^{a_1}$ & $\tilde{Y}_{t, B}^{a_2}$  & $\cdots$ & $\tilde{Y}_{t, B}^{a_K}$ \\
%   \bottomrule
% \end{tabular}
% \end{table}

\section{Concluding Discussion}
Here, we outline a novel end-to-end contextual bandit framework for personalised discount code allocation in e-commerce. Unlike traditional supervised learning methods, DISCO addresses the challenges posed by partial information and data sparsity, by employing an action encoding scheme that enables shared learning across similar actions, and using Thompson sampling to manage the inherent trade-off between exploration and exploitation. We demonstrate the ability of our framework to support both high predictive accuracy in extrapolation (via information sharing), as well the expected monotonicity between discount depths and subsequent purchasing (negative price elasticity). Additionally, we embed our predictive model within a constrained integer program, which affords us a high degree of operational control, and demonstrate that the overall algorithm is still able to efficiently learn and improve over time.  

The methods employed here outline an efficient and performant framework for employing active learning techniques within a practical setting, and can be used in many product areas to take algorithmic actions that balance exploration and exploitation. DISCO exhibits high data efficiency by leveraging Bayesian log-linear regression: despite the high variance in customer behaviour, this approach requires information from only two previous discount campaigns to yield accurate predictions, and the framework is able to generate performant context representations from customer data that is easily obtained through standard business operations. The proposed bandit framework can potentially be applied to a variety of personalisation problems, such as product recommendations or targeting in customer relationship management (CRM). For example, in product recommendations, the framework could be used to dynamically suggest items to users based on their previous interactions and preferences. By training customer and item embeddings based on their context and interactions, these embeddings can be used in a Bayesian log-linear regression (similar to Eq. 6) to facilitate exploration. Similarly, in CRM, the framework could help in identifying the best time and channel to reach out to customers, tailoring messages to their specific needs and past behavior. 
The constant time complexity ensures scalability to millions of customers or users, making it an attractive solution for large-scale applications. Given these promising applications, we save these ideas for future research.

\begin{credits}
\subsubsection{\ackname} We’d like to thank Roger Howard, Priscilla Fearn, Lakshmi Mukkawar, Naveen Kumar Kaveti, Ashton Hills, and Michael Neely for their support with this project, and Jacob Lang and C. H. Bryan Liu for their helpful discussions.
\end{credits}
%% Note that this preceding line implies that you store your BibTeX references in a file called 'mybibliography.bib'. If you instead store your references in a file with a different name, for instance 'references.bib', the preceding line should read '\bibliography{references}'. Whatever you do, DO NOT put the file name extension .bib inside the \bibliography command; this will trip up LaTeX compilers. 
%
% If you do not want to use BibTeX, you can also type up the bibliography exactly as you see fit, using the following structure:

\end{document}